\documentclass{article}
\usepackage{color,soul} 

\usepackage[final,nonatbib]{nips_2017}

\usepackage[utf8]{inputenc} 
\usepackage[T1]{fontenc}    
\usepackage{hyperref}       
\usepackage{url}            
\usepackage{booktabs}       
\usepackage{amsfonts}       
\usepackage{nicefrac}       
\usepackage{microtype}      

\usepackage{amsmath}
\usepackage[pdftex]{graphicx}
\graphicspath{ {./images/} } 
\usepackage{subcaption}

\usepackage{todonotes}

\title{Learning from lions: inferring the utility of agents from their trajectories}

\author{
  Adam D. Cobb \\
  Department of Engineering Science\\
  University of Oxford\\
  \texttt{acobb@robots.ox.ac.uk} \\
   \And
   Andrew Markham \\
   Department of Computer Science \\
   University of Oxford \\
   \texttt{andrew.markham@cs.ox.ac.uk} \\
   \And
  Stephen J. Roberts \\
  Department of Engineering Science\\
  University of Oxford\\
  \texttt{sjrob@robots.ox.ac.uk} \\ \\
}

\begin{document}

\maketitle

\begin{abstract}
We build a model using Gaussian processes to infer a spatio-temporal vector field from observed agent trajectories. Significant landmarks or influence points in agent surroundings are jointly derived through vector calculus operations that indicate presence of sources and sinks. We evaluate these influence points by using the Kullback--Leibler divergence between the posterior and prior Laplacian of the inferred spatio-temporal vector field.
Through locating significant features that influence trajectories, our model aims to give greater insight into underlying causal utility functions that determine agent decision-making.
A key feature of our model is that it infers a joint Gaussian process over the observed trajectories, the time-varying vector field of utility and canonical vector calculus operators.
We apply our model to both synthetic data and lion GPS data collected at the Bubye Valley Conservancy in southern Zimbabwe.

\end{abstract}

\section{Introduction}
Inferring the possible cause of an agent's behaviour from observing its actions is an important area of research across multiple domains. Generic solutions to this problem typically exploit Inverse Reinforcement Learning (IRL) \cite{russell1998learning} and include learning preference value functions from choices \cite{chu2005preference}, as well as utility, value or policy surfaces from observed actions in a space. In this paper we consider observed trajectories, influenced by a time- and space-varying utility function, and look to infer the spatio-temporal utility function from observed tracks. Our work is motivated by a desire to further understand animal interaction with an environment, fuelled by the existence of long-term GPS data. Instead of making deductions about likely tracks, given our knowledge of features in the environment, we look at inverting the task by inferring these significant features in the landscape from tracks.
Our model constructs Gaussian processes (GPs) for jointly inferring a spatio-temporal vector field and canonical vector calculus operations, allowing us to estimate a time-varying map of sources and sinks of potential influence.

Representing the utility function using a GP model is applied to IRL applications in both \cite{levine2011nonlinear,qiao2011inverse}, which improve on previous IRL techniques that constrain the reward to be a linear combination of features \cite{abbeel2004apprenticeship,ramachandran2007bayesian,ziebart2008maximum}.
Instead of treating the problem of inferring characteristics of the utility function through finding an optimal policy that matches observed agent demonstrations, we take advantage of using a GP model to infer the distributions over trajectories and their joint derivatives. These distributions provide an estimate of significant attractors and repellers in an agent's surrounding which influence the observed agent's dynamics.

The use of GPs to model vector fields can be seen in \cite{wahlstrom2013modeling} to directly model magnetic fields. They use divergence-free and curl-free kernels to enforce linear constraints on their models. In \cite{jidling2017linearly}, a method for incorporating any linear constraint is presented. To the best of our knowledge, apart from the above work on modeling vector fields directly with Gaussian processes, the derivative properties of GPs are not currently exploited to infer vector calculus operations from lower order observations.

Previous techniques for studying how animals interact with the environment rely on using GPS data to build density maps to construct probability distributions of their likely locations \cite{horne2007analyzing,laver2008critical}. Although more recent approaches have incorporated time \cite{lyons2013home} into these models, current methods do not focus on inferring the driving force behind animal actions. This work looks to address this gap in the literature, providing not only a tool to aid in animal behaviour studies, but also as an addition to the growing body of work in inverse reinforcement learning. 

The rest of the paper is organised as follows: Section \ref{sec:prelim} introduces the GP and its associated derivative manipulations for vector calculus.
The model is discussed in Section \ref{sec:model}, which is followed by a description of the synthetic data experiments in Section \ref{sec:exp}. The results of applying the model to the lion GPS data are presented in Section \ref{sec:results}. Finally, we discuss the implications of our results and comment on future work in Section \ref{sec:conclusion}.

\section{Preliminaries}
\label{sec:prelim}

As a requirement for our model, we introduce the Gaussian process, defined by its mean function $\boldsymbol{\mu}(\mathbf{x})$ and covariance function $\mathbf{K}(\mathbf{x,x'})$ \cite{rasmussen2006gaussian}.
The mean and covariance functions are parameterized by their hyperparameters and encode prior information into the model, such as smoothness, periodicity and any known underlying behaviour.
In our work, $\boldsymbol{\mu}(\mathbf{x})$ is set to zero as our data is preprocessed to have a zero mean. We define a function, distributed via a GP, as follows: 
\begin{equation}\label{eq:GP}
\mathbf{f(x)} \sim \mathcal{GP}\left(\boldsymbol{\mu}(\mathbf{x}),\mathbf{K}(\mathbf{x,x'})\right).
\end{equation}
Manipulating Gaussian identities using Bayes' rule gives formulae for the GP's posterior mean,
\begin{equation}\label{eq:mean_GP}
\mathbb{E}\left[f(\mathbf{x}_*)\right] = \mathbf{k}_{\mathbf{x}_*\mathbf{x}}^{\top}(\mathbf{K}_{\mathbf{x}\mathbf{x}}+\sigma^2\mathbf{I})^{-1} \mathbf{y},
\end{equation}
and posterior covariance,
\begin{equation}\label{eq:var_GP}
\mathbb{V}\left[f(\mathbf{x}_*)\right] = {k}_{\mathbf{x}_*\mathbf{x}_*} - \mathbf{k}_{\mathbf{x}_*\mathbf{x}}^{\top}(\mathbf{K}_{\mathbf{x}\mathbf{x}}+\sigma^2\mathbf{I})^{-1}\mathbf{k}_{\mathbf{x}_*\mathbf{x}},
\end{equation}
where $\mathbf{x_*}$ is a test point under question and $\sigma^2$ is the noise variance hyperparameter.

Any affine transformation of Gaussian distributed variables remain jointly Gaussian distributed.
As differentiation is an affine operation, applying this property to any collection of random variables distributed by a GP, gives jointly Gaussian distributed derivatives, $\mathbf{f'(x)}$. 
For a test point $\mathbf{x_*}$ and corresponding output $f(\mathbf{x}_*)$, the derivatives associated with the GP in Equation \eqref{eq:GP} are distributed with posterior mean,
\begin{equation}\label{eq:GP_der_mean}
\mathbb{E}\left[\frac{\partial^{n} f(\mathbf{x}_*)}{\partial\mathbf{x}_*^{n}} \right] = \frac{\partial^{n} \mathbf{k}_{\mathbf{x}_*\mathbf{x}}^{\top}}{\partial\mathbf{x}_*^{n}}(\mathbf{K}_{\mathbf{x}\mathbf{x}}+\sigma^2\mathbf{I})^{-1} \mathbf{y},
\end{equation}
and posterior covariance,
\begin{equation}\label{eq:GP_der_var}
\mathbb{V}\left[\frac{\partial^{n} f(\mathbf{x}_*)}{\partial\mathbf{x}_*^{n}} \right] = \frac{\partial^{2n} {k}_{\mathbf{x}_*\mathbf{x'}_*}}{\partial\mathbf{x}_*^{n} \partial\mathbf{x'}_*^{n}} - \frac{\partial^{n} \mathbf{k}_{\mathbf{x}_*\mathbf{x}}^{\top}}{\partial\mathbf{x}_*^{n}}(\mathbf{K}_{\mathbf{x}\mathbf{x}}+\sigma^2\mathbf{I})^{-1}\frac{\partial^{n} \mathbf{k}_{\mathbf{x}_*\mathbf{x}}}{\partial\mathbf{x}_*^{n}}.
\end{equation}
We define Equation \eqref{eq:GP_der_mean} as the predictive mean of the $n^{\text{th}}$ derivative with respect to any test points $\mathbf{x}_*$ and Equation \eqref{eq:GP_der_var} as its corresponding variance.

The choice of covariance selected throughout this paper is the squared exponential kernel,
\begin{equation}\label{eq:Kse}
k_{\text{SE}}(\mathbf{x}_* , \mathbf{x}_i) = l^2 \exp{\left(-\frac{1}{2}(\mathbf{x}_* - \mathbf{x}_i)^{\top}\boldsymbol{\Lambda}^{-1} (\mathbf{x}_* - \mathbf{x}_i)\right)},
\end{equation}
with $\mathbf{x}_*$ and $\mathbf{x}_i$ corresponding to a test point and training point respectively.
The hyperparameter  $\boldsymbol{\Lambda}$, is a diagonal matrix of input scale lengths, where each element determines the relevance of its corresponding dimension and the output scale, denoted by $l^2$, controls the magnitude of the output \cite{roberts2013gaussian}. Although the choice of kernel is somewhat arbitrary, our choice is motivated by the desire to obtain smooth measures over arbitrary derivative functions and the ease by which the kernel can be differentiated and used in combination with Equations \eqref{eq:GP_der_mean} and \eqref{eq:GP_der_var}.
The following formulae define the squared exponential kernel for the first and second order derivatives \cite{mchutchon2013differentiating}:
\begin{equation}\label{eq:Kse_1_der}
\frac{\partial k_{\text{SE}}(\mathbf{x}_*,\mathbf{x}_i)}{\partial \mathbf{x}_*} = - \boldsymbol{\Lambda}^{-1} (\mathbf{x}_* - \mathbf{x}_i) k_{\text{SE}}(\mathbf{x}_*,\mathbf{x}_i)
\end{equation}
\begin{equation}\label{eq:Kse_2_der}
\frac{\partial^2 k_{\text{SE}}(\mathbf{x}_*,\mathbf{x}_i)}{\partial \mathbf{x}_*^2} =  \boldsymbol{\Lambda}^{-1}\left( (\mathbf{x}_* - \mathbf{x}_i)(\mathbf{x}_* - \mathbf{x}_i)^\top\boldsymbol{\Lambda}^{-1} - \mathbf{I}\right) k_{\text{SE}}(\mathbf{x}_*,\mathbf{x}_i).
\end{equation}
We note at this point that estimating derivatives using a joint GP model over the function and derivatives \cite{brook2016emission,holsclaw2013gaussian} offers a benign noise escalation in comparison to numerical differentiation.

\subsection{Vector calculus with GPs}
\label{sec:vec_calc}

We define each dimension of the vector field, $\mathbf{V}_t = \left[V_x\ V_y\right]^{\top}_t$, at time $t$ to be modelled by a multi-input, multi-output GP with a three-dimensional input tuple consisting of $\boldsymbol{\mathcal{X}} = (x,y,t)$, being the spatial and time components of our trajectory.
This GP is constructed by introducing a separable kernel \cite{alvarez2012kernels}, such that
$$
\left[
\begin{array}{c}
V_x\\
V_y
\end{array}
\right]
\sim \mathcal{GP}\left(
\left[
\begin{array}{c}
\mu_x\\
\mu_y
\end{array}
\right],
\left[
\begin{array}{cc}
k_x(\boldsymbol{\mathcal{X}},\boldsymbol{\mathcal{X}}') & 0 \\
0 & k_y(\boldsymbol{\mathcal{X}},\boldsymbol{\mathcal{X}}')
\end{array}
\right]\right)
$$
contains an independently learned kernel for each output dimension.
Applying Equation \eqref{eq:GP_der_mean} to this GP model, by jointly inferring the derivatives in the $x$ and $y$ directions, gives the time dependent posterior for each random variable in the following tuple:
$$
\left( V_x,V_y,\frac{\partial V_x}{\partial x},\frac{\partial V_y}{\partial y}, \frac{\partial V_x}{\partial y},\frac{\partial V_y}{\partial x}\right)_t.
$$

We combine these predictive derivatives using Equations \eqref{eq:div} and \eqref{eq:curl} to infer probability distributions over the divergence and curl of the vector field $\mathbf{V}$:
\begin{equation}\label{eq:div}
\boldsymbol{\nabla}\cdot \mathbf{V} = \frac{\partial V_x}{\partial x} + \frac{\partial V_y}{\partial y},
\end{equation}
\begin{equation}\label{eq:curl}
\boldsymbol{\nabla}\times \mathbf{V} = \left| 
\begin{array}{ccc}
\mathbf{i} & \mathbf{j} & \mathbf{k} \\
\frac{\partial}{\partial x} & \frac{\partial}{\partial y} & \frac{\partial}{\partial z} \\
V_x & V_y & 0 
\end{array} \right| =  \mathbf{k}\left(\frac{\partial V_y}{\partial x} - \frac{\partial V_x}{\partial y}\right).
\end{equation}
By simple application of the appropriate operators, we may readily define the time-varying spatial Laplacian:
\begin{equation}\label{eq:laplace}
\boldsymbol{\nabla} \cdot (\boldsymbol{\nabla} \psi) \equiv \boldsymbol{\nabla}^2 \psi,
\end{equation}
where $\psi$ is a scalar potential function. This (time-varying) Laplacian is of key importance as it defines \emph{sources} and \emph{sinks} in the spatial domain.

\section{Model}\label{sec:model}
Our model builds upon the theory introduced in section \ref{sec:prelim}.
The objective is to design a model that can indicate influential features in an agent's landscape from observing the trajectories.

A trajectory $\boldsymbol{\zeta}_a$ for agent $a$ is defined as a collection of timestamped locations $\mathbf{x} \in \mathcal{R}^2$ or tuples $(\mathbf{x},t)$.
The elements of the vector $\mathbf{x}$ are referred to as $x$ and $y$ for the purposes of this model.
We also make the assumption that each agent acts according to a utility or potential function $\phi(\mathbf{x},{t})$, which is dependent on time and space.
Whilst interacting with the environment, each agent aims to maximise this utility function at all times.

\subsection{Fitting to the agent trajectories}\label{sec:traj_fit}

The first stage of the model uses Equations \eqref{eq:mean_GP} and \eqref{eq:var_GP} to fit a GP to each trajectory $\boldsymbol{\zeta}_a$.
Using a single GP with a separable kernel, Equation \eqref{eq:joint_GP1} defines our GP prior, which is a joint distribution over the path and its derivatives with respect to time, where we denote $\ddot{f}_x$ and $\ddot{f}_y$ as the second-order time derivatives in the $x$ and $y$ directions:
\begin{equation}\label{eq:joint_GP1}
p(\vec{\mathbf{f}},\ddot{f}_x,\ddot{f}_y) = \mathcal{GP}\left(\mathbf{0},\left[\begin{array}{cc}
k_x(t,t') & 0\\
0 & k_y(t,t')
\end{array}\right]\right).
\end{equation}
For each of the $x$ and $y$ components of $\vec{\mathbf{f}}$, a set of hyperparameters are learned for the separable kernel.
The input space of this GP model is time $t$ and the output space consists of the $x$ and $y$ components of $\vec{\mathbf{f}}$. If an agent trajectory, $\boldsymbol{\zeta}_a$, consists of $N$ data points, we can use the posterior, $p(\vec{\mathbf{f}},\ddot{f}_x,\ddot{f}_y \mid \boldsymbol{\zeta}_a)$, to infer expectations for the derivatives at each of the $n$ data points.

Second-order derivatives are inferred at this stage, as we make the tacit assumption that an agent acts in accordance with a second-order dynamical system, i.e. the agent obeys Newtonian dynamics. This assumption means that an agent's underlying utility induces a ``force''  of influence on the agent, thus generating an acceleration (we here take the `influence-mass' of the agent as unity). More formally, this induces an acceleration equal to the derivative of the agent utility:
$$
\ddot{f}_x = \frac{\partial \phi}{\partial {x}},\ \ \ddot{f}_y = \frac{\partial \phi}{\partial {y}}.
$$

When dealing with a multi-agent system of $M$ homogeneous agents, a trajectory model can be calculated for each agent to form the set of joint distributions,
$$\Big\{p(\vec{\mathbf{f}},\ddot{f}_x,\ddot{f}_y \mid \boldsymbol{\zeta}_a)\Big\}_{a=0}^M.$$
From this GP model we are able to jointly predict the velocity and acceleration at any point on an agent's trajectory for $M$ agents.
The next layer of our model combines these to construct a probability distribution over the extended agent environment.

\subsection{Inferring the vector field and Laplacian}\label{sec:inf_VF}
In order to infer the gradient of the utility function, $\boldsymbol{\nabla} \mathbf{\phi(\mathbf{x}},t)$, the set of inferred second derivatives for all $M$ agents is propagated through a second GP, which also has a separable kernel model, as below:
\begin{multline} \label{eq:joint_GP2}
p\Big(\vec{\mathbf{\mathbf{V}}}(\mathbf{x},t),\frac{\partial V_x}{\partial x},\frac{\partial V_y}{\partial y}, \frac{\partial V_x}{\partial y},\frac{\partial V_y}{\partial x} \mid \{\ddot{f}_x,\ddot{f}_y,\boldsymbol{\zeta}_a\}_{a=0}^M\Big)\\ = \mathcal{GP}\left(\left[
\begin{array}{c}
\boldsymbol{\mu}_x^{\text{post}} \\
\boldsymbol{\mu}_y^{\text{post}}
\end{array}\right]
,\left[\begin{array}{cc}
k_x^{\text{post}}(\boldsymbol{\zeta}_a,\boldsymbol{\zeta}_a') & 0\\
0 & k_y^{\text{post}}(\boldsymbol{\zeta}_a,\boldsymbol{\zeta}_a')
\end{array}\right]\right).
\end{multline}
The vector $\vec{\mathbf{\mathbf{V}}}(\mathbf{x},t) = \left[{V}_x\ {V}_y\right]^{\top}$ consists of two random variables that model the acceleration in the two axes and the superscript label `$\text{post}$' refers to the calculated posterior mean and covariance. 
Equation \eqref{eq:joint_GP2} combines the $M$ multiple agent paths into one model and enables predictions to be made at different points in space that are not constrained to a single agent trajectory as in Equation \eqref{eq:joint_GP1}. The input-output pairs for this GP model are the $x,y$ and $t$ values in each $\boldsymbol{\zeta}_a$ that correspond to the $\ddot{f}_x$ and $\ddot{f}_y$ values.

The Newtonian assumption made in Section \ref{sec:traj_fit} is formally included as
$$
\boldsymbol{\nabla} \mathbf{\phi(\mathbf{x}},t) \propto \vec{\mathbf{\mathbf{V}}}(\mathbf{x},t).
$$
The partial derivatives from Equation \eqref{eq:joint_GP2} can thence be combined to calculate a distribution of the divergence of $\vec{\mathbf{\mathbf{V}}}$. It follows that this divergence is proportional to the Laplacian under the same assumption,
$$
\boldsymbol{\nabla}^2 \mathbf{\phi(\mathbf{x}},t) \propto \boldsymbol{\nabla}\cdot \vec{\mathbf{\mathbf{V}}}(\mathbf{x},t).
$$
In particular, our interest lies in the estimation of the Laplacian of the utility function, as it indicates \emph{sources} and \emph{sinks} of the potential function in the environment. In this context, we regard \emph{sinks} as agent \emph{attractors} and \emph{sources} as agent \emph{repellers}.

\subsection{Metric for locating significant attractors and repellers: Kullback--Leibler divergence} 
Our metric of change from prior field to posterior field is measured via the Kullback-Leibler (KL) divergence \cite{kullback1951information}. The motivation for selecting the KL divergence comes from its ability measure a distance between two distributions by taking into account both the mean and variance. This provides a natural indication of the informativeness of spatial locations, at given times, and in the context of our application, offers a measure of \emph{trajectory-influencing locations}.

Given the model at time $t$, each point in space has an associated potential field distribution, defined via the GP posterior as a univariate normal distribution.
The KL divergence can be readily calculated as the difference between two univariate normal distributions, namely the prior and posterior \cite{duchi2007derivations}, as below:
\begin{equation}\label{eq:KL_div}
D_{\text{KL}}(p_{\text{prior}} \mid \mid p_{\text{posterior}}) = \frac{1}{2}\left( \frac{\sigma_{\text{pr}}^2}{\sigma_{\text{po}}^2} + \frac{(\mu_{\text{po}}-\mu_{\text{pr}})^2}{\sigma_{\text{po}}^2} - 1 + \ln\left(\frac{\sigma_{\text{po}}}{\sigma_{\text{pr}}}\right)\right),
\end{equation}
where
$$
p_{\text{prior}} = \mathcal{N}(\mu_{\text{pr}},\sigma_{\text{pr}}^2), \ \ p_{\text{posterior}} = \mathcal{N}(\mu_{\text{po}},\sigma_{\text{po}}^2).
$$

We refer back to Equation \eqref{eq:div} and \eqref{eq:laplace} in order to calculate the following prior at location $\boldsymbol{\mathcal{X}}$ in space-time:
$$
 \left(\frac{\partial V_x}{\partial x}+\frac{\partial V_y}{\partial y}\right) \sim \mathcal{N}\Big(0,\frac{\partial^2 k_x(\boldsymbol{\mathcal{X}},\boldsymbol{\mathcal{X}}')}{\partial x^2}+\frac{\partial^2 k_y(\boldsymbol{\mathcal{X}},\boldsymbol{\mathcal{X}}')}{\partial y^2}\Big),
$$
where, $
\frac{\partial^2 k_x(\boldsymbol{\mathcal{X}},\boldsymbol{\mathcal{X}}')}{\partial x^2}+\frac{\partial^2 k_y(\boldsymbol{\mathcal{X}},\boldsymbol{\mathcal{X}}')}{\partial y^2} =
\frac{h_x^2}{\lambda_x^2}+\frac{h_y^2}{\lambda_y^2}.
$ The hyperparameters $h_x$ and $\lambda_x$ are the output and input scale lengths of the $x$-part of the separable kernel in the GP model, with $h_y$ and $\lambda_y$ corresponding to the $y$-part.

As our interest lies in determining attractors and repellers in the field, a further addition to the KL divergence in Equation \eqref{eq:KL_div} is to multiply it by the sign of the posterior mean of the Laplacian. This multiplication carries over the prediction of negative sinks and positive sources, whilst measuring the divergence from the zero prior mean. We refer to this trivial extension as the \emph{signed KL divergence}.

\subsection{Hyperparameter optimisation} Optimisation of the hyperparameters is achieved through maximisation of the GPs marginal log-likelihood using scaled conjugate gradient approaches. We initialise hyperparameters for spatial models with domain knowledge, where available. In our example, lions are dynamic objects with maximal accelerations, which places natural priors on the relationships between spatial movement and time variation.

\section{Application to synthetic data}\label{sec:exp}
We apply our model to synthetic data where the true utility function and its derivatives are known, so as to test the performance of the approach.
The experiment consists of a multi-agent system of four identical agents, whose dynamics we observe. Our goal is to infer, from trajectories alone, the underlying potential value function.

\subsection{Utility function}
The utility function for the test example is defined as a two-component Gaussian mixture model, where each Gaussian has a time-varying covariance that changes harmonically according to:
\begin{equation}\label{eq:utility}
\phi(\mathbf{x},t) = \mathcal{N}\big(\boldsymbol{\mu}_1, (\sin t + a)\mathbf{I}\ \big) + \mathcal{N}\big(\boldsymbol{\mu}_2, (\cos t + b)\mathbf{I}\ \big),
\end{equation}
where $a$ and $b$ are constants and $\boldsymbol{\mu}_1$, $\boldsymbol{\mu}_2$ are the distribution means. We also denote $\mathbf{I}$ as the identity matrix. 
The first and second order derivatives are calculated using the known derivatives of a multivariate normal distribution $p(\mathbf{x})$ \cite{petersen2008matrix}:
\[
\frac{\partial p(\mathbf{x})}{\partial \mathbf{x}}=-p(\mathbf{x})\boldsymbol{\Sigma}^{-1}(\mathbf{x-m}),
\]
\[
\frac{\partial^2 p(\mathbf{x})}{\partial \mathbf{x} \partial \mathbf{x}^\top}=p(\mathbf{x})\Big(\boldsymbol{\Sigma}^{-1}(\mathbf{x-m})(\mathbf{x-m})^\top \boldsymbol{\Sigma}^{-1} - \boldsymbol{\Sigma}^{-1}\Big).
\]
These derivatives are then used in conjunction with the vector calculus operations in Equations \eqref{eq:div} and \eqref{eq:laplace} to calculate the true Laplacian of the utility function.

\subsection{Agent model}
Agents are modelled according to a second order dynamical system, whereby, at each time step $t$, the acceleration $\mathbf{\ddot{x}}$, velocity $\mathbf{\dot{x}}$ and position $\mathbf{x}$ are given by:
$$\begin{array}{lcl}
\mathbf{\ddot{x}}_{t+1} & = & \boldsymbol{\nabla} \phi(\mathbf{x}_t,t) + \mathbf{n},\\
\mathbf{\dot{x}}_{t+1} & = & \mathbf{\dot{x}}_t + \eta\ \mathbf{\ddot{x}}_{t+1},\\
\mathbf{x}_{t+1} & = & \mathbf{x}_t + \eta\ \mathbf{\dot{x}}_{t+1}.
\end{array}$$
The random variable, $\mathbf{n}$, corresponds to additive noise distributed according to a two-dimensional multivariate Gaussian and the variable $\eta$ is the update increment.

\subsection{Experimental results}\label{sec:syn_data}
\begin{figure}[h]
    \centering
    \begin{subfigure}[b]{1.\textwidth}
        \includegraphics[width=\textwidth]{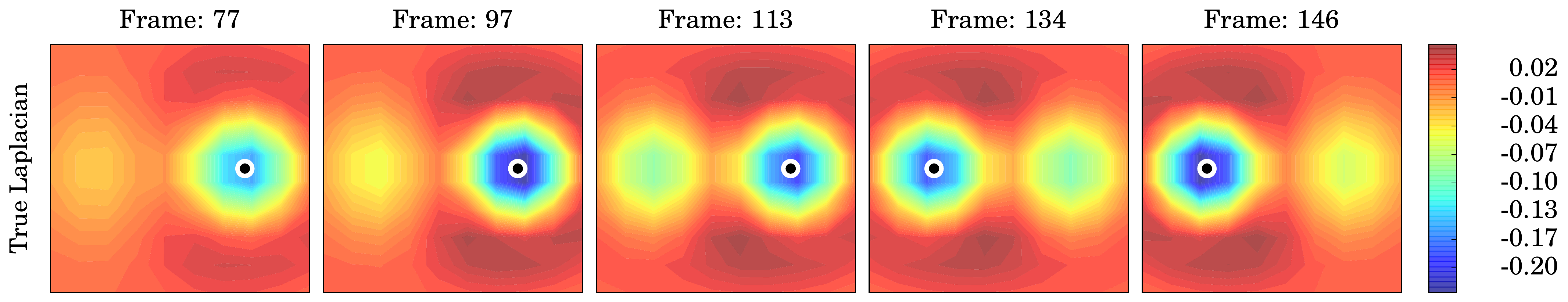}
        \label{skl_top}
    \end{subfigure}\\[-3ex]
     \begin{subfigure}[b]{1.\textwidth}
        \includegraphics[width=\textwidth]{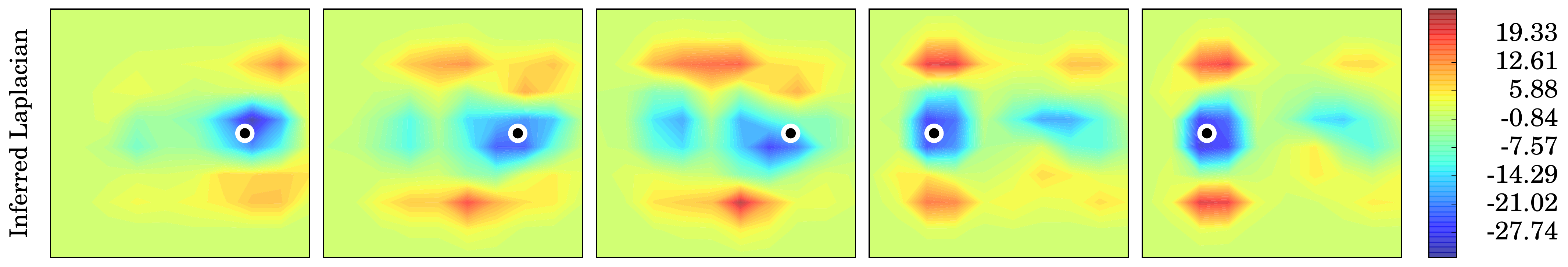}
        \label{skl_mid}
    \end{subfigure}\\[-3ex]
    \begin{subfigure}[b]{1.\textwidth}
        \includegraphics[width=\textwidth]{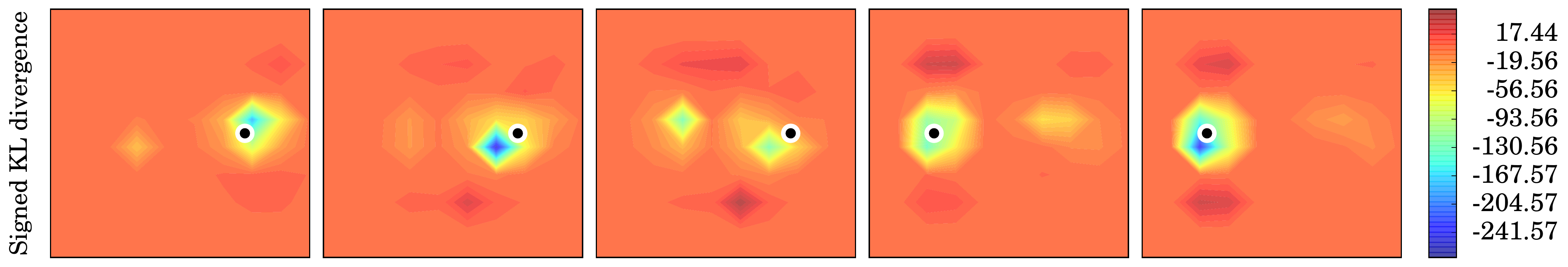}
        \label{skl_bot}
    \end{subfigure}
    \caption{Top row: Laplacian of true utility function. Middle row: inferred Laplacian of utility function. Bottom row: signed KL divergence. The location of the global minimum of the true Laplacian is indicated via the black and white marker. The predicted locations of the sinks, given by both the signed KL divergence and the posterior Laplacian, align well with their true locations.}\label{fig:frames_figures}
\end{figure}

Four agents were initialised at locations $[2,1],[2,-1],[-2,1]$ and $[-2,-1]$ with a velocity of zero. The parameters of the utility function in Equation \eqref{eq:utility} were $a=b=2.1$ and $\boldsymbol{\mu}_1=[-1.5,0]$, $\boldsymbol{\mu}_2 = [1.5,0]$.
The model stepped through $200$ time-steps and the inferred Laplacian of the utility function was calculated from the four agent trajectories in $\{\boldsymbol{\zeta}_a\}_{a=1}^4$.
The subset of frames displayed in Figure \ref{fig:frames_figures} show how the attractors of the true utility functions are tracked across time by both the posterior inferred Laplacian and the signed KL divergence. The uncertainty in our model is displayed through the signed KL divergence.
The black and white markers denote the location of the dominant attractor, at that time step, given by the ground truth in the first row.
For all three rows, regions of attraction are indicated by blue and it can be seen that the model is able to track the true attractors across time.

\begin{figure}[h]
    \centering
        \includegraphics[width=1.0\textwidth]{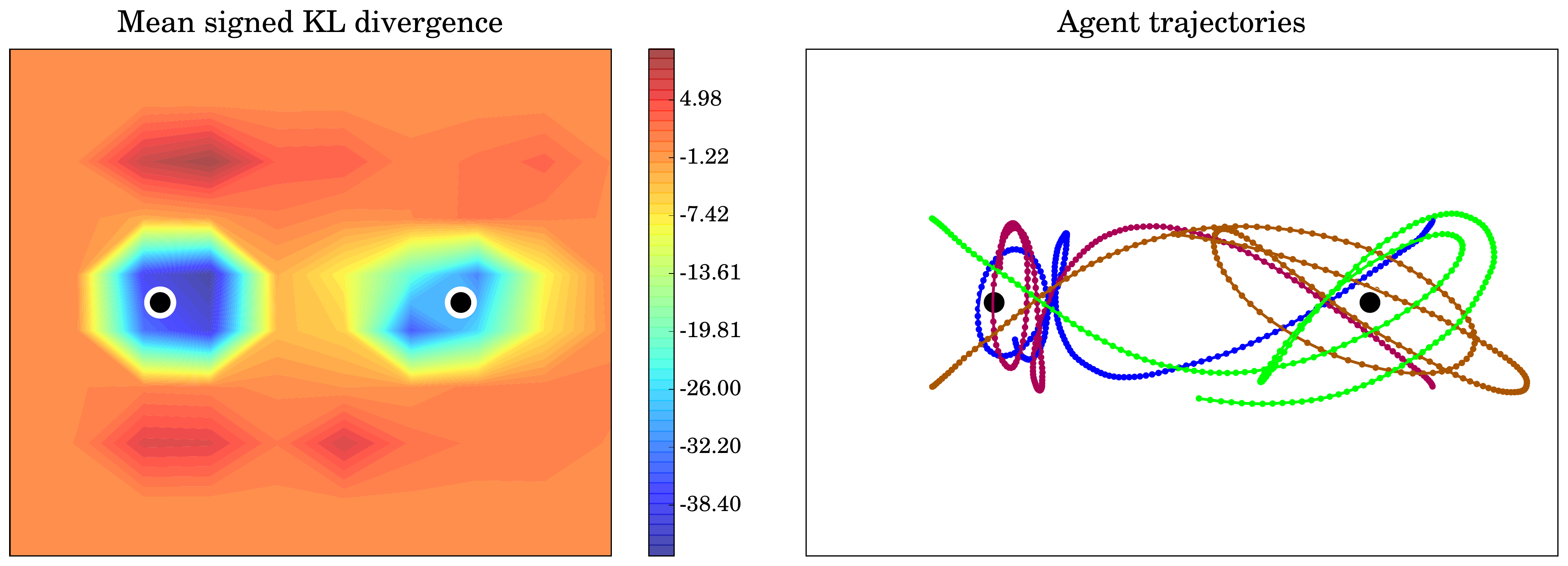}
        \caption{The mean signed KL divergence across time is shown in the left plot and the corresponding agent trajectories are shown in the right plot. Each agent path is denoted via a separate color. The true locations of the attractors are indicated by the black markers. Taking the average over the $200$ frames of data allows us to see the mean inferred position of attractors in the agents' field, which coincide with the known positions.}
        \label{fig:skl_mean_4agents}
\end{figure}

In Figure \ref{fig:skl_mean_4agents}, the left-hand plot displays the mean KL divergence taken over time, showing the emergence of the average location and size of the attractors. The right-hand plot shows each agent's path and how their trajectories are drawn to the sink locations. This figure demonstrates that our model has correctly inferred the sink locations without prior knowledge of their location.

\section{Lion trajectories}\label{sec:results}

In this section, we examine how our model can be used to determine drivers in the landscape that impact an animal's behaviour.
Previous techniques for studying how animals interact with the environment rely on using GPS data to build density maps to construct probability distributions of their likely locations \cite{horne2007analyzing,laver2008critical}.
Although more recent approaches have incorporated time \cite{lyons2013home} into these models, current methods do not focus on inferring the driving force behind animal actions and instead simply show where an animal is likely to be found.
Therefore we apply our model to a subset of lion GPS data to show a possible way of inferring the location of influential features in a lion's landscape.  This perspective differs from traditional ecological techniques, which rely on knowledge of the environment to test a hypothesis.
In particular, our focus is to build from work by Valeix et al. (2010) \cite{valeix2010key} that show waterholes are key attractive features on lions' trajectories, using knowledge of waterhole locations.
Inverting this perspective, applying our model to this data set aims to infer locations of waterholes that are important to a particular lion, without supplying this as a prior. 

\begin{figure}[h]
    \centering
    \begin{subfigure}[b]{1.\textwidth}
        \includegraphics[width=\textwidth]{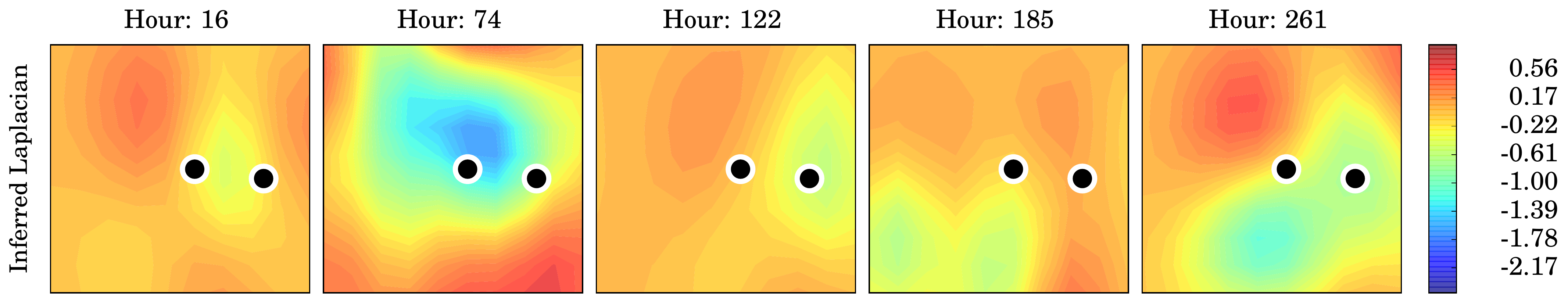}
        \label{lion_top}
    \end{subfigure}\\[-3ex]
     \begin{subfigure}[b]{1.\textwidth}
        \includegraphics[width=\textwidth]{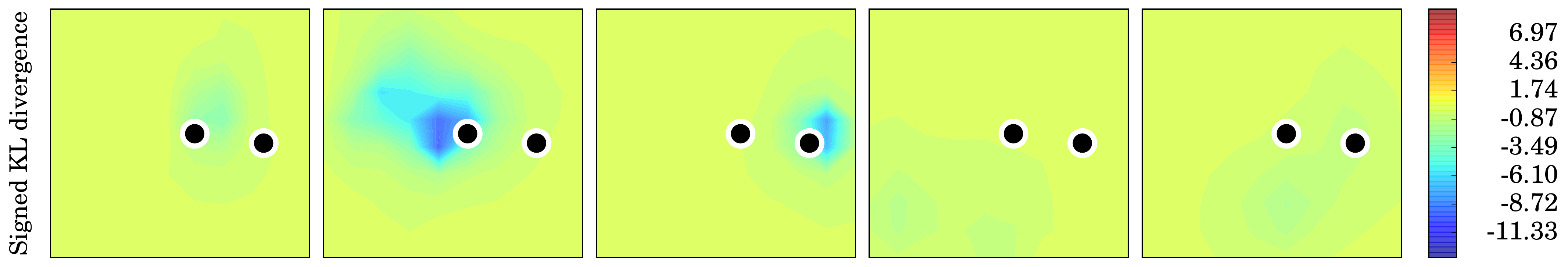}
        \label{lion_bot}
    \end{subfigure}\\[-3ex]
    \caption{Top row: inferred Laplacian of utility function. Bottom row: signed KL divergence. Each column represents the model at a given hour since the start of the subset of lion data. The relative location of the waterholes are denoted by the black and white markers. Blue regions indicate attractive regions. The signed KL divergence infers attractors occuring at different times that coincide with  the waterholes, without prior knowledge of the waterhole locations.}\label{fig:frames_figures_lion}
\end{figure}

\begin{figure}[h]
    \centering
        \includegraphics[width=1.0\textwidth]{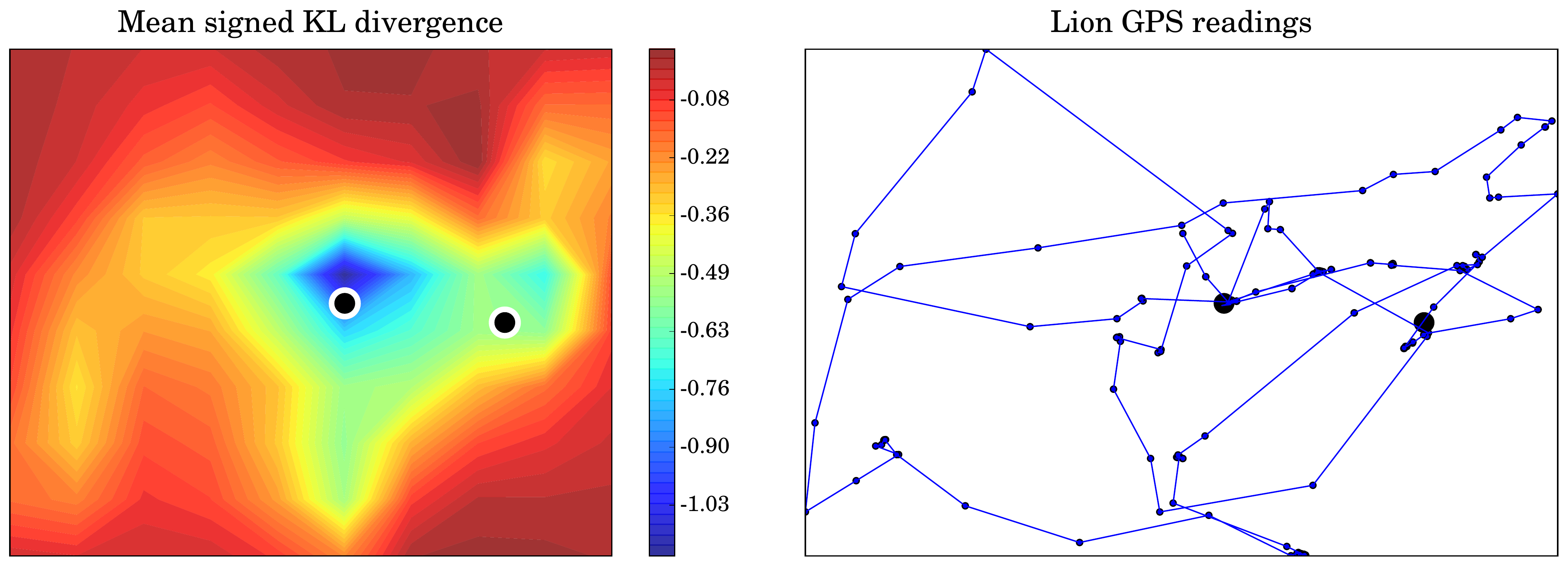}
        \caption{The left plot displays the mean of the signed KL divergence across time to give an average estimate of the lion attractor locations. The blue areas indicate large negative values corresponding to the sinks. The right-hand plot contains the lion trajectory data. The black markers give the relative locations of waterholes, provided by the GIS data. This data set covers $200$ GPS readings that were collected over $11$ days and $16$ hours. The spacial scale for this figure is $15 \times 15$ km$^2$. }
        \label{fig:lion_figures}
\end{figure}

The model is applied to GPS data collected from lions living at the Bubye Valley Conservancy (BVC) in southern Zimbabwe. Figures \ref{fig:frames_figures_lion} and \ref{fig:lion_figures} display the results of applying the model to $200$ data points covering $11$ days and $16$ hours of a particular lion's GPS readings.
The $x$ and $y$ locations have been normalised in keeping with the relative geographic information system (GIS) data.
In both figures the locations of waterholes are included to show their correspondence with predicted regions of attraction.
Figure \ref{fig:frames_figures_lion} shows the model inferring attractor locations at different time steps, where the units are in hours since the start of the data set.
These regions can be seen to overlap the known waterhole locations at certain time steps.
In Figure \ref{fig:lion_figures} the left-hand plot superimposes the GIS data of waterholes on the mean signed KL divergence produced from the lion trajectory, where the mean is taken across time.
The corresponding lion trajectory GPS readings are shown in the right-hand plot.
The mean signed KL divergence indicates that there are two waterholes located in areas near local minima, leading to the reasonable assumption that they may behave as influential attractive features. In this way, we can infer decision-making strategies behind observed trajectories. Richer models could investigate further effects such as location of prey and other lions in the ecosystem.


\section{Conclusion}\label{sec:conclusion}
Through applying layers of Gaussian Processes combined with vector calculus, the model provides a useful way of inferring significant features in an agent's surroundings that influence its behaviour. We demonstrate the results of applying this model to a synthetic data set, where the true attractors are known.
Our model successfully finds the attractors that determine the agents' trajectories.
Furthermore, applying the model to a subset of lion GPS data allows us to determine a spatio-temporal influence measure which highlights the two waterholes that the lion visits most frequently. Unlike previous work in this domain,  such information was not provided to the model. Furthermore, unlike density estimation based methods, our model makes predictions that depend on dynamic decisions made by the lion.

Further work will extend this model to propagate uncertainty from the GPS readings through to the final Laplacian predictions, adapting techniques used in \cite{damianou2013deep,girard2003gaussian}. We aim to extend our model to considerably larger data sets through utilising sparse GP models as needed \cite{titsias2009variational}.

\subsubsection*{Acknowledgements}

We would like to thank the shareholders of Bubye Valley Conservancy for allowing us to conduct field research on the conservancy and the management staff for assistance in the field. Adam Cobb is sponsored by the AIMS CDT (\url{http://aims.robots.ox.ac.uk}) and the EPSRC (\url{https://www.epsrc.ac.uk}).

\small

\bibliographystyle{unsrt}
\bibliography{bibliography_nips}

\begin{thebibliography}{10}

\bibitem{russell1998learning}
Stuart Russell.
\newblock Learning agents for uncertain environments.
\newblock In {\em Proceedings of the eleventh annual conference on
  Computational learning theory}, pages 101--103. ACM, 1998.

\bibitem{chu2005preference}
Wei Chu and Zoubin Ghahramani.
\newblock Preference {L}earning with {G}aussian {P}rocesses.
\newblock In {\em Proceedings of the 22nd international conference on Machine
  learning}, pages 137--144. ACM, 2005.

\bibitem{levine2011nonlinear}
Sergey Levine, Zoran Popovic, and Vladlen Koltun.
\newblock Nonlinear {I}nverse {R}einforcement {L}earning with {G}aussian
  processes.
\newblock In {\em Advances in Neural Information Processing Systems}, pages
  19--27, 2011.

\bibitem{qiao2011inverse}
Qifeng Qiao and Peter~A Beling.
\newblock Inverse {R}einforcement {L}earning with gaussian {P}rocess.
\newblock In {\em American Control Conference (ACC), 2011}, pages 113--118.
  IEEE, 2011.

\bibitem{abbeel2004apprenticeship}
Pieter Abbeel and Andrew~Y Ng.
\newblock Apprenticeship {L}earning via {I}nverse {R}einforcement {L}earning.
\newblock In {\em Proceedings of the twenty-first international conference on
  Machine learning}, page~1. ACM, 2004.

\bibitem{ramachandran2007bayesian}
Deepak Ramachandran and Eyal Amir.
\newblock Bayesian {I}nverse {R}einforcement {L}earning.
\newblock {\em Urbana}, 51(61801):1--4, 2007.

\bibitem{ziebart2008maximum}
Brian~D Ziebart, Andrew~L Maas, J~Andrew Bagnell, and Anind~K Dey.
\newblock Maximum {E}ntropy {I}nverse {R}einforcement {L}earning.
\newblock In {\em AAAI}, volume~8, pages 1433--1438. Chicago, IL, USA, 2008.

\bibitem{wahlstrom2013modeling}
Niklas Wahlstr{\"o}m, Manon Kok, Thomas~B Sch{\"o}n, and Fredrik Gustafsson.
\newblock Modeling {M}agnetic {F}ields {U}sing {G}aussian {P}rocesses.
\newblock In {\em Acoustics, Speech and Signal Processing (ICASSP), 2013 IEEE
  International Conference on}, pages 3522--3526. IEEE, 2013.

\bibitem{jidling2017linearly}
Carl Jidling, Niklas Wahlstr{\"o}m, Adrian Wills, and Thomas~B Sch{\"o}n.
\newblock Linearly constrained {G}aussian processes.
\newblock {\em arXiv preprint arXiv:1703.00787}, 2017.

\bibitem{horne2007analyzing}
Jon~S Horne, Edward~O Garton, Stephen~M Krone, and Jesse~S Lewis.
\newblock Analyzing animal movements using {B}rownian bridges.
\newblock {\em Ecology}, 88(9):2354--2363, 2007.

\bibitem{laver2008critical}
Peter~N Laver and Marcella~J Kelly.
\newblock A {C}ritical {R}eview of {H}ome {R}ange {S}tudies.
\newblock {\em Journal of Wildlife Management}, 72(1):290--298, 2008.

\bibitem{lyons2013home}
Andrew~J Lyons, Wendy~C Turner, and Wayne~M Getz.
\newblock Home range plus: a space-time characterization of movement over real
  landscapes.
\newblock {\em Movement Ecology}, 1(1):2, 2013.

\bibitem{rasmussen2006gaussian}
Carl~Edward Rasmussen.
\newblock Gaussian {P}rocesses for {M}achine {L}earning.
\newblock 2006.

\bibitem{roberts2013gaussian}
Stephen Roberts, M~Osborne, M~Ebden, Steven Reece, N~Gibson, and S~Aigrain.
\newblock Gaussian processes for time-series modelling.
\newblock {\em Phil. Trans. R. Soc. A}, 371(1984):20110550, 2013.

\bibitem{mchutchon2013differentiating}
Andrew McHutchon.
\newblock Differentiating {G}aussian {P}rocesses.
\newblock \url{http://mlg.eng.cam.ac.uk/mchutchon/DifferentiatingGPs.pdf},
  2013.

\bibitem{brook2016emission}
PR~Brook, A~Karastergiou, S~Johnston, M~Kerr, RM~Shannon, and SJ~Roberts.
\newblock Emission-rotation correlation in pulsars: new discoveries with
  optimal techniques.
\newblock {\em Monthly Notices of the Royal Astronomical Society},
  456(2):1374--1393, 2016.

\bibitem{holsclaw2013gaussian}
Tracy Holsclaw, Bruno Sans{\'o}, Herbert~KH Lee, Katrin Heitmann, Salman Habib,
  David Higdon, and Ujjaini Alam.
\newblock Gaussian {P}rocess {M}odeling of {D}erivative {C}urves.
\newblock {\em Technometrics}, 55(1):57--67, 2013.

\bibitem{alvarez2012kernels}
Mauricio~A Alvarez, Lorenzo Rosasco, Neil~D Lawrence, et~al.
\newblock Kernels for {V}ector-{V}alued {F}unctions: a {R}eview.
\newblock {\em Foundations and Trends{\textregistered} in Machine Learning},
  4(3):195--266, 2012.

\bibitem{kullback1951information}
Solomon Kullback and Richard~A Leibler.
\newblock On {I}nformation and {S}ufficiency.
\newblock {\em The annals of mathematical statistics}, 22(1):79--86, 1951.

\bibitem{duchi2007derivations}
John Duchi.
\newblock Derivations for {L}inear {A}lgebra and {O}ptimization.
\newblock {\em Berkeley, California}, 2007.

\bibitem{petersen2008matrix}
Kaare~Brandt Petersen, Michael~Syskind Pedersen, et~al.
\newblock The {M}atrix {C}ookbook.
\newblock {\em Technical University of Denmark}, 7:15, 2008.

\bibitem{valeix2010key}
Marion Valeix, Andrew~J Loveridge, Zeke Davidson, Hillary Madzikanda, Herv{\'e}
  Fritz, and David~W Macdonald.
\newblock How key habitat features influence large terrestrial carnivore
  movements: waterholes and {A}frican lions in a semi-arid savanna of
  north-western {Z}imbabwe.
\newblock {\em Landscape Ecology}, 25(3):337--351, 2010.

\bibitem{damianou2013deep}
Andreas Damianou and Neil Lawrence.
\newblock Deep {G}aussian {P}rocesses.
\newblock In {\em Artificial Intelligence and Statistics}, pages 207--215,
  2013.

\bibitem{girard2003gaussian}
Agathe Girard, Carl~Edward Rasmussen, Joaquin~Quinonero Candela, and Roderick
  Murray-Smith.
\newblock Gaussian {P}rocess {P}riors {W}ith {U}ncertain {I}nputs {A}pplication
  to {M}ultiple-{S}tep {A}head {T}ime {S}eries {F}orecasting.
\newblock {\em Advances in neural information processing systems}, pages
  545--552, 2003.

\bibitem{titsias2009variational}
Michalis~K Titsias.
\newblock Variational {L}earning of {I}nducing {V}ariables in {S}parse
  {G}aussian {P}rocesses.
\newblock In {\em AISTATS}, volume~5, pages 567--574, 2009.

\end{thebibliography}

\end{document}